\documentclass[11pt,letterpaper,DIV=19,abstract=true]{scrartcl}
\usepackage[utf8]{inputenc}
\usepackage[linesnumbered, ruled, vlined]{algorithm2e}
\usepackage{amssymb,amsmath,amsthm,amsfonts}
\usepackage{bm}
\usepackage[x11names]{xcolor}
\usepackage[backend=biber,style=alphabetic,sorting=nyt]{biblatex}
\usepackage{booktabs}
\usepackage{caption}
\usepackage{circuitikz}
\usepackage{subcaption}
\usepackage{courier}
\usepackage{enumitem}
\usepackage{graphicx}
\usepackage{listings}
\usepackage{mathrsfs}
\usepackage{mathtools}
\usepackage{multirow}
\usepackage{svg}
\usepackage{xurl}
\usepackage{hyperref}

\DeclarePairedDelimiter\parens{\lparen}{\rparen}

\DeclareMathOperator*{\argmax}{arg\,max}

\newcommand{\func}[2]{#1\parens*{#2}}

\title{Extending Spike-Timing Dependent Plasticity to Learning Synaptic Delays}
\date{}
\author{$^1$Marissa Dominijanni, $^2$Alexander Ororbia, $^1$Kenneth W. Regan\\$^1$State University of New York at Buffalo \\
$^2$Rochester Institute of Technology}

\begin{document}
\maketitle

\abstract{\noindent\begin{addmargin*}[4em]{4em}Synaptic delays play a crucial role in biological neuronal networks, where their modulation has been observed in mammalian learning processes. In the realm of neuromorphic computing, although spiking neural networks (SNNs) aim to emulate biology more closely than traditional artificial neural networks do, synaptic delays are rarely incorporated into their simulation. We introduce a novel learning rule for simultaneously learning synaptic connection strengths and delays, by extending spike-timing dependent plasticity (STDP), a Hebbian method commonly used for learning synaptic weights. We validate our approach by extending a widely-used SNN model for classification trained with unsupervised learning. Then we demonstrate the effectiveness of our new method by comparing it against another existing methods for co-learning synaptic weights and delays as well as against STDP without synaptic delays. Results demonstrate that our proposed method consistently achieves superior performance across a variety of test scenarios. Furthermore, our experimental results yield insight into the interplay between synaptic efficacy and delay.\end{addmargin*}}

\section{Introduction}
\label{sec:intro}

Traditional artificial neural networks (ANNs) are fundamentally trained with only a single \emph{kind} of parameter, one that controls the relative impact for each input on a given output. This type of parameter can be further subdivided into two categories: \emph{weights} and \emph{biases}. The biases control the output of artificial neurons in the absence of nonzero input whereas the weights scale the impact of each input on the output. Equation \ref{eq:ann-linear} demonstrates the typical calculation employed by ANNs to compute the value of an output $\mathbf{y}_j$ given an input vector $\mathbf{x}$, parameterized by weights $\mathbf{W}$, biases $\mathbf{b}$, and nonlinearity $f$:
\begin{equation}
    \mathbf{y}_j = \func{f}{\sum_i \mathbf{W}_{ji} \, \mathbf{x}_i + \mathbf{b}_j} . \label{eq:ann-linear}
\end{equation}

In contrast to ANNs, the atom of information in spiking neural networks (SNNs) is no longer a real-valued number but instead a \emph{spike train}. In continuous time, this train is represented as a series of timestamps where spike pulses occur. For discrete time simulations, we instead represent this computationally as a vector of zeros and ones, where each element is a time step and spikes occur at nonzero elements. Spike trains constitute a core signaling mechanism by which biological neurons communicate over long distances. From an information and machine learning perspective, the increase in signal dimensionality provides a new degree of freedom to explore for parameterizing neural networks. Equation \ref{eq:snn-linear} shows the key calculation used by SNNs to compute the output of a neuron $\mathbf{y}_j$ at time $t$ from input signals $\mathbf{x}$ parameterized by weights $\mathbf{W}$, delays $\mathbf{D}$, and biases $\mathbf{b}$. Here, $f_j$ and $g_{ji}$ are stateful functions representing the neuronal dynamics and synaptic kinetics, respectively. Although included here, the bias term is often excluded from the parameterization of SNNs.
\begin{equation}\label{eq:snn-linear}
    \mathbf{y}_j(t) = \func{f_j}{\sum_i \mathbf{W}_{ji} \, g_{ji}(\mathbf{x}_i(t - \mathbf{D}_{ji})) + \mathbf{b}_j}
\end{equation}
Popular designs of $f_j$ include models of cellular membrane potential, such as the leaky integrate-and-fire (LIF) process and variants thereof.
The delay in signaling between neurons is itself a direct consequence of the brain being a physical, event-driven system. Biologically, there are two major sources for this delay:
1) the time required for a signal to travel from the cell body along the axon of the ``sending'' neuron (i.e., the \emph{axonal delay}), and,
2) the time required for a signal to travel from the dendritic spine to the cell body along the dendritic tree of the ``receiving'' neuron (i.e., the \emph{dendritic delay}) \cite{manor_effect_1991,boudkkazi_release-dependent_2007}. Together, these two sources represent the \emph{synaptic delay} through which we seek to parameterize an SNN.

This work is motivated by the notion that synaptic delays are a powerful tool to leverage the capacity of SNNs. Not only has it been observed that heterogeneous synaptic delays are present in connections between neurons \cite{manor_effect_1991}, and that the precise timing of spikes plays an important role in synaptic plasticity \cite{bi_synaptic_1998}, but also that synaptic delays themselves are modulated during learning \cite{lin_modulation_2002,boudkkazi_release-dependent_2007}. For instance, synaptic delays inform the localization of sound by the mammalian auditory pathway \cite{glackin_spiking_2010}. In addition to their importance in biological learning systems, synaptic delays offer significant representational power in machine learning. For instance, Grappolini and Subramoney showed \cite{grappolini_beyond_2023} that an SNN without synaptic delays performs similarly to an SNN with trainable synaptic delays and randomly initialized weights, when trained with backpropagation of errors \cite{rumelhart_learning_1986} using a surrogate gradient method. They inferred that connection strengths and delays each provide comparable representational capacity.

In this study, we present a biologically-plausible method for training the synaptic weights of spiking neural networks. We extend this method to the general adjustment of synaptic delays and provide a theoretical comparison to a key related effort. Finally, we compare all three methods by modifying an existing SNN model to incorporate dynamic, learnable synaptic delays.

\section{Methodology}
\label{sec:methods}

\subsection{Background}
\label{ssec:background}

\begin{figure}[!ht]
    \centering
    \includegraphics{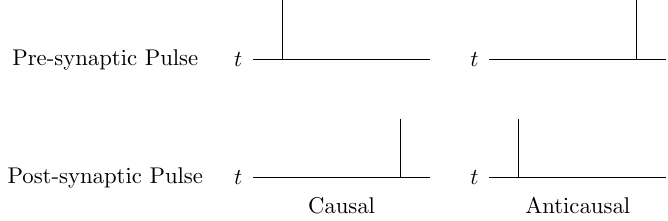}
    \caption{The causal relationship of spike pairs between a pre-synaptic and post-synaptic neuron.}
    \label{fig:spike-pair}
\end{figure}

Unlike ANNs, which are typically fitted to data via backpropagation of errors (backprop) \cite{rumelhart_learning_1986}, SNNs in computational neuroscience are more commonly trained using biologically-plausible plasticity rules \cite{ororbia_brain-inspired_2023}, often based on the relative timing of pre-synaptic and post-synaptic spike trains. A commonly used method is spike-timing dependent plasticity (STDP) \cite{kistler_modeling_2000}, which instantiates Hebbian learning, well-known under the mantra: ``neurons that fire together, wire together'' \cite{hebb_organization_1949} (however, the timing matters \cite{markram_history_2011}). More precisely, Hebbian learning only potentiates connections where the pre-synaptic neuron took part in the firing of the post-synaptic neuron; STDP considers specifically pairs of pre-synaptic and post-synaptic spikes between connected neurons. If the pre-synaptic spike occurs prior to a post-synaptic spike, then the ordering is considered to be causal in nature (the pre-synaptic pulse gave rise to the post-synaptic pulse) and the synaptic connection is potentiated. If the post-synaptic spike occurs prior to the pre-synaptic spike, then the ordering is considered to be anticausal and the synaptic connection is depressed. See Figure \ref{fig:spike-pair} for a visual representation of the potential causal and anticausal relationships that two spikes can have.

\begin{figure}[!t]
    \centering
    \includegraphics{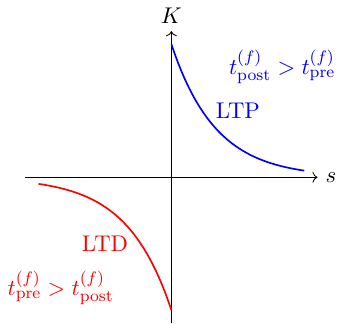}
    \caption{Plot of long-term potentiation (LTP), in blue, and long-term depression (LTD), in red, induced by an exponential kernel configured for STDP learning.}
    \label{fig:exp-stdp}
\end{figure}

In order to define the update rule for STDP given the above orderings that spikes might have, we need a way to mathematically represent spike trains. Equation \ref{eq:nrf} gives the neural response function and is utilized to represent spikes $\mathcal{F}$ relative to a time $t$ as the summation of Dirac delta terms in integral equations:
\begin{equation}\label{eq:nrf}
    S(t) = \sum_{f \in \mathcal{F}} \func{\delta}{t - t^{(f)}} .
\end{equation}
Given the above definition of a spike train, we next turn to the STDP update rule for a given connection weight $w$:
\begin{equation}\label{eq:stdp}
    \begin{split}
        \left(\frac{dw}{dt}\right)_{\text{post},\text{pre}}^\text{STDP} = \alpha &+
        S_\text{post}(t) \left[\beta_\text{post}
        + \int_{0}^{\infty} K_+(s) S_\text{pre}(t - s) ds\right] \\
        &+ S_\text{pre}(t) \left[\beta_\text{pre}
        + \int_{0}^{\infty} K_-(s) S_\text{post}(t - s) ds\right]
    \end{split}
\end{equation}
where $\alpha$ is the baseline weight update, $\beta_\text{post}$ and $\beta_\text{pre}$ are non-Hebbian terms controlling the application of an update at the point of occurrence of a post-synaptic or pre-synaptic spike, respectively. $K_+$ and $K_-$ are the kernels used for convolving the pre-synaptic and post-synaptic spike trains up until the current time $t$, respectively; the convolution itself is performed based on the offset $s$ to the present \cite{kistler_modeling_2000}. The above equation results in a sparse update rule when excluding the non-Hebbian terms, where updates are gated by spikes and are proportional to convolved spike trains. Furthermore, note that a commonly used kernel used for this update is the exponential kernel:
\begin{equation}\label{eq:exp-kernel}
    K_\pm(s) = A_\pm \exp \left(\frac{-s}{\tau_\pm}\right)
\end{equation}
which implies that each spike provides an instantaneous response with amplitude $A_\pm$, which decays exponentially with time constant $\tau_\pm$. Figure \ref{fig:exp-stdp} plots the impact of a spike from this kernel over time.

When using STDP for online learning, typically the integral equation for STDP is not used but instead a local variable called a \emph{spike trace} is maintained and used to compute the requisite convolution. Formally, the resulting traces are:
\begin{align}
    \frac{dX}{dt} &= -\frac{X}{\tau} + S(t)A \label{eq:cum-trace} \\
    \frac{dX}{dt} &= -\frac{X}{\tau} + S(t) \left[A - \frac{X}{k}\right] \label{eq:sat-trace}
\end{align}
where Equation \ref{eq:cum-trace} defines the \emph{cumulative trace}
and \ref{eq:sat-trace} defines the \emph{saturating trace}, respectively \cite{morrison_phenomenological_2008}. The former of these is equivalent to using the exponential kernel whereas the latter caps the trace such that the amplitude from $k$ spikes (excluding decay and the decreasing contributions due to convex combination) causes it to max out. When $k = 1$, only the most recent spike has an effect and this is referred to as the \emph{nearest trace}.

Using spike traces, the formulation for STDP is often simplified to:
\begin{equation}\label{eq:trace-stdp}
    \left(\frac{dw}{dt}\right)_{\text{post},\text{pre}}^\text{STDP} = S_\text{post}(t) X_\text{pre}(t) + S_\text{pre}(t) X_\text{post}(t)
\end{equation}
where the spike traces can take any form of a local variable that represents the convolved spike train and the non-Hebbian terms, i.e., $\alpha$, $\beta_\text{post}$, and $\beta_\text{pre}$, are dropped.

\subsection{Delay-Shifted Spike-Timing Dependent Plasticity}
\label{ssec:delay-stdp}

In order to extend the STDP paradigm to training synaptic delays, we first leverage the observation that potentiating synaptic connections corresponds to \emph{increasing} weight values and \emph{decreasing} delay values \cite{nadafian_bioplausible_2024}. Intuitively, the length of time it takes for a pre-synaptic signal to be received by a post-synaptic neuron can be thought of as inversely proportional to the ``strength'' of its connection (or its efficacy). To this end, we introduce delay-shifted spike-timing dependent plasticity (DS-STDP).

Below, we present the equations for DS-STDP updates for both the synaptic weights and the synaptic delays.

Concretely, the update rule for a synapse's weight is:
\begin{equation}\label{eq:ds-stdp-weight}
    \begin{split}
        \left(\frac{dw}{dt}\right)^\text{DS-STDP}_{\text{post},\text{pre}}
        = \alpha &+
        S_\text{post}(t) \left[\beta_\text{post}
        + \int_{0}^{\infty} K_+(s) S_\text{pre}(t - d - s) ds\right] \\
        &+ S_\text{pre}(t - d) \left[\beta_\text{pre}
        + \int_{0}^{\infty} K_-(s) S_\text{post}(t - s) ds\right]
    \end{split}
\end{equation}
while the update for the synapse's delay is as follows:
\begin{equation}\label{eq:ds-stdp-delay}
    \begin{split}
        \left(\frac{dd}{dt}\right)^\text{DS-STDP}_{\text{post},\text{pre}}
        = \alpha^\prime &+
        S_\text{post}(t) \left[\beta^\prime_\text{post}
        + \int_{0}^{\infty} K^\prime_-(s) S_\text{pre}(t - d - s) ds\right] \\
        &+ S_\text{pre}(t - d) \left[\beta^\prime_\text{pre}
        + \int_{0}^{\infty} K^\prime_+(s) S_\text{post}(t - s) ds\right]
    \end{split}
\end{equation}

Note that the above very closely follow the normal STDP update rules with two notable exceptions. Firstly, the pre-synaptic spikes are shifted by a learned delay $d$, corresponding to when they are observed by the post-synaptic neuron. Secondly, the potentiative and depressive kernels for delay learning, $K^\prime_+$ and $K^\prime_-$, are flipped in position from their weight learning counterparts.

As with STDP, spike traces can be used in place of the convolved spike trains. The reformulated rules for the weight and delay updates with DS-STDP then become the following---for the weights, we obtain:
\begin{equation}\label{eq:trace-ds-stdp-weight}
    \left(\frac{dw}{dt}\right)^\text{DS-STDP}_{\text{post},\text{pre}} = S_\text{post}(t) X_\text{pre}(t - d) + S_\text{pre}(t - d) X_\text{post}(t) ,
\end{equation}
and for the delays, we arrive at:
\begin{equation}\label{eq:trace-ds-stdp-delay}
    \left(\frac{dd}{dt}\right)^\text{DS-STDP}_{\text{post},\text{pre}} = S_\text{post}(t) X^\prime_\text{pre}(t - d) + S_\text{pre}(t - d) X^\prime_\text{post}(t) .
\end{equation}

Note that, because of the delay shift applied to the pre-synaptic spike train and the resulting spike trace, a buffer of prior activity bounded by the maximum permitted value of $d$ must also be stored.

\subsection{Comparison and Analysis}
\begin{figure}[!ht]
    \centering
    \includegraphics{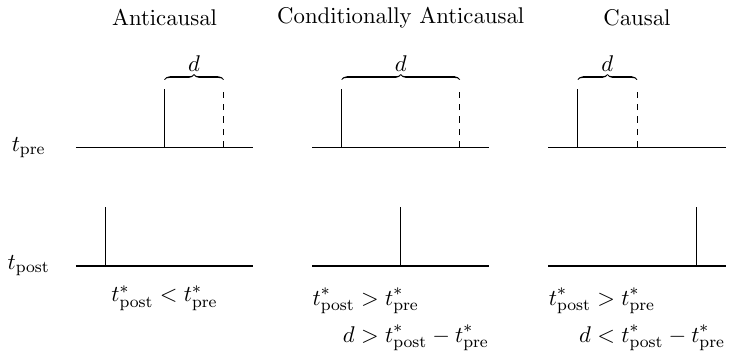}
    \caption{Possible configurations of a pre-synaptic and post-synaptic spike pair when including synaptic delays.}
    \label{fig:delayed-pair}
\end{figure}

A recent innovation in unsupervised delay learning, delay-related spike-timing dependent plasticity (DR-STDP) offers the closest point of comparison to our proposed DS-STDP \cite{nadafian_bioplausible_2024}. Rather than using convolved spike trains as the basis for computing updates, this scheme considers the difference in time between the most recent spike pair when factoring in the synaptic delay. Specifically, DR-STDP requires a formulation that specifies the time of the last spike (time-of-last-spike), up to and including time $t$, as follows:
\begin{equation}\label{eq:last-spike-time}
    t^\ast(t) = \max \left\{t^\prime \in [0, t] \mid S(t^\prime) \neq 0 \right\}.
\end{equation}

DR-STDP then defines the difference in time between when the last pre-synaptic spike was \emph{observed} and the last post-synaptic spike was \emph{generated}. This difference is formally defined as:
\begin{equation}\label{eq:last-spike-diff}
    \begin{split}
        t_\Delta(t)
        &= t^\ast_\text{post}(t) - t^\ast_\text{pre}(t) - d(t) \\
        &= \tau^\ast_\text{pre}(t) - d(t) -  \tau^\ast_\text{post}(t) \quad \text{where } \tau^\ast(t) = t - t^\ast(t) .
    \end{split}
\end{equation}

Finally, DR-STDP \cite{nadafian_bioplausible_2024} prescribes the synaptic weight updates to be:
\begin{equation}\label{eq:dr-stdp-weight}
    \big(\Delta w\big)_{\text{post},\text{pre}}^\text{DR-STDP} =
    \begin{cases}
        A_+ \exp(-\lvert t_\Delta(t) \rvert / \tau_+) &\text{if } t_\Delta(t) \geq 0 \\
        A_- \exp(-\lvert t_\Delta(t) \rvert / \tau_-) &\text{if } t_\Delta(t) < 0
    \end{cases}
\end{equation}
and the synaptic delay updates to be:
\begin{equation}\label{eq:dr-stdp-delay}
    \big(\Delta d\big)_{\text{post},\text{pre}}^\text{DR-STDP} =
    \begin{cases}
        A^\prime_- \exp(-\lvert t_\Delta(t) \rvert / \tau^\prime_-) &\text{if } t_\Delta(t) \geq 0 \\
        A^\prime_+ \exp(-\lvert t_\Delta(t) \rvert / \tau^\prime_+) &\text{if } t_\Delta(t) < 0
    \end{cases} .
\end{equation}

Note that these update terms are directly based on the exponential kernel, and correspond to $K_\pm(\lvert t_\Delta(t) \rvert)$ for weight updates and $K_\mp(\lvert t_\Delta(t) \rvert)$ for delay updates.

In order to compare the update behavior of DS-STDP and DR-STDP, we will first consider the initial update produced by the three possible configurations of spike pairs, given in Figure \ref{fig:delayed-pair}. For this, we will assume Hebbian update parameters for both methods and an exponential kernel for DS-STDP. Since this only considers a pair of spikes, the results would be the same whether using either a cumulative or saturating trace. By evaluating each possible case, we find that the initial updates for DS-STDP and DR-STDP are identical in value, but occur at different times: all cases are enumerated in Table \ref{tbl:ds-dr-stdp-init}. Specifically, since DR-STDP only considers the time of spike generation whereas DS-STDP is anchored relative to when a spike is received, anticausal and conditionally anticausal updates are shifted by $d$ in time.

\begin{table}[htbp]
    \centering
    \caption{Comparison of initial synaptic delay updates for a single spike pair between the proposed DS-STDP using an exponential kernel and DR-STDP.}
    \label{tbl:ds-dr-stdp-init}
    \begin{tabular}{cclc}
        \toprule
        Condition & Update & Time & Rule \\
        \midrule
        \multirow{2}{*}{$t^\ast_\text{post} < t^\ast_\text{pre} \leq t^\ast_\text{pre} + d$} & \multirow{2}{*}{$A^\prime_+\exp\left(\frac{t^\ast_\text{post} - t^\ast_\text{pre} - d}{\tau^\prime_+}\right)$}
        & $t^\ast_\text{pre} + d$ & DS-STDP\\
        && $t^\ast_\text{pre}$ & DR-STDP \\
        \midrule
        \multirow{2}{*}{$ t^\ast_\text{pre} < t^\ast_\text{post} < t^\ast_\text{pre} + d$} & \multirow{2}{*}{$A^\prime_+\exp\left(\frac{t^\ast_\text{post} - t^\ast_\text{pre} - d}{\tau^\prime_+}\right)$}
        & $t^\ast_\text{pre} + d$ & DS-STDP\\
        && $t^\ast_\text{post}$ & DR-STDP \\
        \midrule
        \multirow{2}{*}{$t^\ast_\text{pre} \leq t^\ast_\text{pre} + d < t^\ast_\text{post}$} & \multirow{2}{*}{$A^\prime_-\exp\left(\frac{t^\ast_\text{pre} + d - t^\ast_\text{post}}{\tau^\prime_-}\right)$}
        & $t^\ast_\text{post}$ & DS-STDP\\
        && $t^\ast_\text{post}$ & DR-STDP \\
        \bottomrule
    \end{tabular}
\end{table}

\begin{figure}[!ht]
    \centering
    \includegraphics{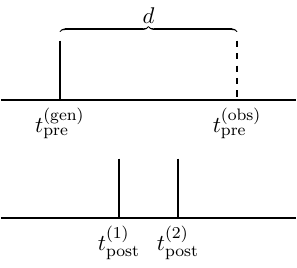}
    \caption{Triplet of one pre-synaptic and two post-synaptic spikes where the delay shifts the configuration from causal to anticausal.}
    \label{fig:delayed-triplet}
\end{figure}

For more complex patterns of spikes, the timing of updates can introduce differences in the initial updates between DS-STDP and DR-STDP. Consider a situation where two post-synaptic spikes occur between when a pre-synaptic spike is generated and when it is observed, a case depicted in Figure \ref{fig:delayed-triplet}.

Below, for DS-STDP and DR-STDP, the former using cumulative traces, we show how to calculate the sum of initial updates by the time the pre-synaptic spike is observed by the post-synaptic neuron, $t^{(\text{obs})}_\text{pre}$. Specifically, for DS-STDP this is:
\begin{equation}\label{eq:ds-stdp-triplet}
    \begin{aligned}
        \func{\Delta d}{t^{(\text{obs})}_\text{pre}}
        &= A^\prime_+\exp\left(\frac{t^{(2)}_\text{post} - t^{(\text{gen})}_\text{pre} - \func{d}{t^{(\text{obs})}_\text{pre}}}{\tau^\prime_+}\right) \\
        &+ A^\prime_+\exp\left(\frac{t^{(1)}_\text{post} - t^{(\text{gen})}_\text{pre} - \func{d}{t^{(\text{obs})}_\text{pre}}}{\tau^\prime_+}\right)
    \end{aligned}
\end{equation}
and for DR-STDP, this is:
\begin{equation}\label{eq:dr-stdp-triplet}
    \begin{aligned}
        \func{\Delta d}{t^{(2)}_\text{post}} + \func{\Delta d}{t^{(1)}_\text{post}}
        &= A^\prime_+\exp\left(\frac{t^{(2)}_\text{post} - t^{(\text{gen})}_\text{pre} - \func{d}{t^{(2)}_\text{post}}}{\tau^\prime_+}\right) \\
        &+ A^\prime_+\exp\left(\frac{t^{(1)}_\text{post} - t^{(\text{gen})}_\text{pre} - \func{d}{t^{(1)}_\text{post}}}{\tau^\prime_+}\right) .
    \end{aligned}
\end{equation}
Since DR-STDP repeatedly applies updates without gating, the value of the delay changes, causing the sum of updates to drift. When DS-STDP is implemented using nearest spike traces instead, the second term in Equation \ref{eq:ds-stdp-triplet} is dropped entirely.

Although DS-STDP updates are generally sparse, unlike DR-STDP updates, repeated applications of an update can be applied under certain conditions. Specifically, this happens as the result of an increase in the delay, causing a previously observed spike to be shifted such that it is observed again. This degenerates quickly in the anti-Hebbian case, where the delay is increased on a post-synaptic spike. An update that increases the delay could shift the pre-synaptic spike such that it is observed after the same post-synaptic spike that initiated the update. This second update, however, would decrease the delay, preventing it from triggering further updates.

The Hebbian case, however, does not degenerate as the anticausal pair \emph{increases} the delay. Equation \ref{eq:ds-stdp-hebbint} defines the total increase in delay from a pre-synaptic spike in the form of a nonlinear integral equation:
\begin{equation}\label{eq:ds-stdp-hebbint}
    d(\mu) = d_0 + \int_{0}^{\mu} x(s) \delta(s - d(s)) ds .
\end{equation}
For compactness, we let $t_0 = t^\ast_\text{pre}$ and $\mu = t - t_0$. Correspondingly, we let $x$ and $d$ define the post-synaptic kernel and synaptic delay shifted such that $x(0) = X_\text{post}(t_0)$ and $d(0) = d(t_0)$. Finally, we assume that there is no spiking activity on the interval $(0, \mu]$. Each change in delay is controlled by $x(s)$ based on the last post-synaptic spike and updates are only applied where $d(s) = s$ since the Dirac delta term acts as a filter.

We can represent the above for a discrete simulation. Concretely, utilize the following:
\begin{equation}\label{eq:ds-stdp-discrete-hebbint}
    d(n) = d_0 + \sum_{s = 0}^{n} x(s \Delta t) \left[s = \left\lceil \frac{d(s)}{\Delta t}\right\rceil\right] ,
\end{equation}
which defines a discrete formulation of the initial value problem evaluated after $n$ time steps, where $\Delta t$ is the simulation step time. Note that we treat a spike at time $t$ as occurring over the range $[t, t + \Delta t)$.

When $x$ is always less than or equal to half of $\Delta t$, then at most two consecutive updates will be applied. Consider the case where $d(s_0)$ is a nonzero integer multiple $s_0$ of $\Delta t$, implying that the pre-synaptic spike is observed at time $t_0 + s_0 \Delta t$. Then, at $s = s_0$, the delay is nudged upwards by some small value $x(s_0 \Delta t)$. In the next term of the sum, i.e., $s = s_1$, the update is triggered again since it has shifted which temporal bin is being checked for a pre-synaptic spike. Only if $x(s_1 \Delta t)$ is greater than $\Delta t - x(s_0 \Delta t)$ could this then result in the same spike being integrated a third time. Therefore, in discrete simulations, the sparsity of updates will be preserved so long as this condition holds. If this condition does not always hold, updates will be applied until $x(s \Delta t)$ drops back below this threshold.

\section{Experimental Results}
\label{sec:experimental-results}

In order to evaluate DS-STDP in comparison to DR-STDP and STDP, we consider the task of classification in the context the Diehl \& Cook SNN model \cite{diehl_unsupervised_2015}. As in the original, this setup entails unsupervised learning methods to train an SNN's parameters, and then a classifier is post-fitted to the output of the trained neural network (with its parameters frozen).

\subsection{Spiking Network Architecture}
\label{ssec:spiking-network-arch}
\begin{figure}[!t]
    \centering
    \includegraphics{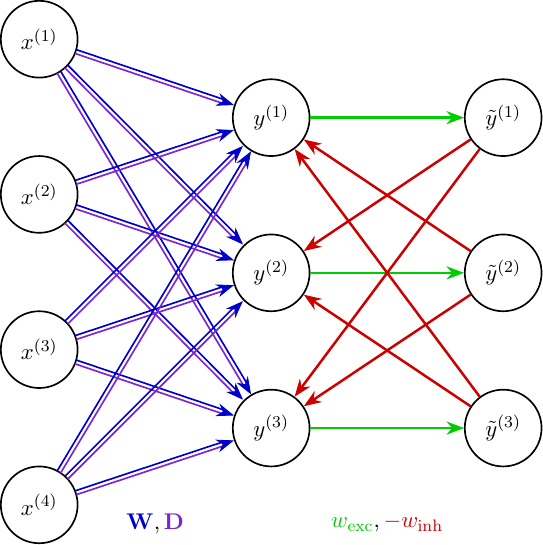}
    \caption{Diehl \& Cook model architecture equipped with synaptic delays.}
    \label{fig:ddc-arch}
\end{figure}

Since two of the unsupervised learning methods under consideration incorporate synaptic delays, we make an alteration to the base model. Figure \ref{fig:ddc-arch} provides a graphical representation of this architecture, with input neurons $x$, excitatory neurons $y$, and inhibitory neurons $\tilde{y}$.

Note that, in this architecture, the number of excitatory neurons $N$ equals the number of inhibitory neurons, and the weights between them are configured as hyperparameters. Each excitatory neuron $y^{(j)}$ has a connection to its corresponding inhibitory neuron $\tilde{y}^{(j)}$ with weight $w_\text{exc}$. Each inhibitory neuron $\tilde{y}^{(j)}$ is connected to every excitatory neuron other than the one that triggered it, i.e., $y^{(k)}$ where $j \neq k$, with weights $-w_\text{inh}$. These are equivalent to linear connections, where the weight matrix of the former is $w_\text{exc} \, \mathbf{I}_N$ (a scalar matrix), and the weight matrix of the latter is $w_\text{inh} \, (\mathbf{I}_N - 1)$ (a  hollow matrix). This mechanism is referred to as \emph{lateral inhibition} and is used to enforce competition among the excitatory neurons.

The particular modification made to the network architecture itself is in the trainable connections between the input and excitatory neurons, where the connections are parameterized by trainable delays $\mathbf{D}$ in addition to trainable weights $\mathbf{W}$. In practice, a buffered set of pre-synaptic spikes and spike traces for a prior time up to $d_\text{max}$ are stored and used to temporally shift the inputs to the network.

\subsection{Classification}
\label{ssec:classification}
Our classification procedure is based on the original rate coding method described in \cite{diehl_unsupervised_2015}. In contrast, we make two modifications: the first is to include an $\ell_1$-normalization step as in \cite{hazan_bindsnet_2018}, and the second is one of our own design, which extends the classification procedure to any score in the domain $[0,1]$ for spike trains where larger values correspond to a stronger response. Specifically, for this modified scheme, in addition to rate coding, we develop a score based on latency coding \cite{johansson_first_2004,gollisch_rapid_2008}.

Formally, in the equations below, we define the \emph{spike rate} and \emph{spike responsiveness} for a neuron $n$ over an interval of time from $t_0$ to $T$. The latter of these is based on latency coding, and is inversely proportional to the latency scaled by the duration of the spike train. The resulting scheme is described by the following:
\begin{align}
    Q_n(t_0) &= \frac{1}{T} [T - \min \{\tau_n(t_0), T\}] \label{eq:spike-responsiveness} \\
    &\text{where } \tau_n = \inf \{t \in [0, T] \mid S_n(t_0 + t) \neq 0\} \nonumber
\end{align}
and rate coding is described by the following:
\begin{equation}\label{eq:spike-rate}
    R_n(t_0) = \frac{1}{T} \int_{0}^{T} S_n(t_0 + t) dt .
\end{equation}
Note that we define the score vector for a neuron $n$ over $K$ classes, where each class $k$ has an associated set of spike trains beginning at times $\mathcal{T}_k$, and $\mathcal{P}_n$ is the scoring function applied to the $n^\text{th}$ neuron. Based on this, we define the class associated with each neuron as follows:
\begin{equation}\label{eq:score-vector}
    \mathbf{v}_n(\mathcal{P}) = \left[\frac{1}{\lvert\mathcal{T}_k\rvert} \sum_{t_s \in \mathcal{T}_k} \mathcal{P}_n(t_s)\right]_{k=1}^{K}
\end{equation}
\begin{equation}\label{eq:association-vector}
    \mathbf{a} = \left[\argmax_k \left[\mathbf{v}_n(\mathcal{P})\right]_k\right]_{n=1}^{N}
\end{equation}
Finally, we define an $N \times K$ matrix where each row specifies to the per-class response for the corresponding neuron:
\begin{equation}\label{eq:classification-matrix}
    \mathbf{C} =
    \begin{bmatrix}
        \left[\left[\mathbf{a}_1 = k\right]\right]_{k=1}^{K} \odot \frac{\mathbf{v}_1(\mathcal{P})}{\lVert \mathbf{v}_1(\mathcal{P}) \rVert_1} \\\\
        \left[\left[\mathbf{a}_2 = k\right]\right]_{k=1}^{K} \odot \frac{\mathbf{v}_2(\mathcal{P})}{\lVert \mathbf{v}_2(\mathcal{P}) \rVert_1} \\\\
        \vdots \\\\
        \left[\left[\mathbf{a}_N = k\right]\right]_{k=1}^{K} \odot \frac{\mathbf{v}_N(\mathcal{P})}{\lVert \mathbf{v}_N(\mathcal{P}) \rVert_1}
    \end{bmatrix} .
\end{equation}
Each neuron only has a nonzero response to the associated class (as though each row were a one-hot label vector), but the response is then scaled by the $\ell_1$-normalized score vector so the degree by which class $k$ ``wins'' is factored into the classification. To compute the logits for a batch of output scores $\mathbf{Y}$, where each row corresponds to a sample and each column responds to the score $\mathcal{P}_n$ for the $n^\text{th}$ output neuron, we use the following:
\begin{equation}\label{eq:logits}
    \mathbf{Z} = \mathbf{Y} \mathbf{C} \oslash \left[\max \left\{1, \sum_{n=1}^{N} \left[\mathbf{a}_n = k\right]\right\}\right]_{k=1}^{K} .
\end{equation}

\subsection{Testing Methodology}
\label{ssec:test-methods}
As with the original Diehl \& Cook paper \cite{diehl_unsupervised_2015}, we performed our classification tests using the standard MNIST dataset \cite{lecun_gradient-based_1998}. The dataset contains 60,000 training and 10,000 testing samples of $28 \times 28$ pixel 8-bit greyscale images representing handwritten digits from $0$ to $9$ along with their corresponding label. From this, the training was split into two subsets---a 50,000 sample class-balanced subset was used as the training set and the remaining 10,000 samples were used as a validation set for hyperparameter tuning.

For conversion of pixel intensities to spike trains, the intensities in the range $\mathbb{Z} \cap [0, 255]$ are normalized to a range $[0, \nu_\text{max}]$ via min-max scaling, where $\nu_\text{max}$ is the maximum expected spike rate. These are then encoded via a homogeneous Poisson process, without refractoriness, into spike trains. This is done by sampling the expected interspike intervals from an exponential distribution, turning them into spike times via cumulative summation, and binning them within the steps of the simulation.

We evaluated models with $100$, $225$, $400$, $625$, and $900$ pairs of excitatory and inhibitory neurons, trained using DS-STDP, DR-STDP, and STDP. In order to compare performance via parameter count as well as neuron count, we evaluated models with $1225$, $1600$, and $2025$ neuron pairs trained using STDP. Experiments were conducted using the same random initialization for each model, with an identical presentation of training batches. Note that some nondeterministic CUDA operations such as the cumulative summation of floating-point values were used, which may introduce small variations between runs and between devices. Hyperparameters are provided for: the general simulation in Table \ref{tbl:general-hyperparams}, for training in Tables \ref{tbl:train-common-hyperparams} and \ref{tbl:train-spec-hyperparameters}, and for neurons and synapses in Table \ref{tbl:neuron-hyperparams}. We remark that the hyperparameters for neurons, synapses, and weight initialization are based on those from the BindsNET implementation of the Diehl \& Cook model \cite{hazan_bindsnet_2018}.

\begin{table}[!b]
    \centering
    \caption{General simulation hyperparameters.}
    \label{tbl:general-hyperparams}
    \begin{tabular}{lcc}
        \toprule
        Hyperparameter & Value \\
        \midrule
        Batch Size, Training ($B$) & $50$ \\
        Simulation Step Length ($\Delta t$) & $1 \text{ ms}$ \\
        Per-Batch Simulation Duration ($T$) & $250 \text{ ms}$ \\
        Maximum Spike Rate ($\nu_\text{max}$) & $127.5 \text{ Hz}$ \\
        \midrule
        Weight Initialization ($\mathbf{W}(t = 0)$) & $\mathcal{U}(0, 0.3)$ \\
        Delay Initialization ($\mathbf{D}(t = 0)$) & $\mathcal{U}(0, 10)$ \\
        \midrule
        Excitatory-to-Inhibitory Weights ($w_\text{exc}$) & $22.5$ \\
        Inhibitory-to-Excitatory Weights ($-w_\text{inh}$) & $-120$ \\
        \bottomrule
    \end{tabular}
\end{table}

\subsubsection{Training and Evaluation}
\label{sssec:train-eval}
In addition to the methods for DS-STDP, DR-STDP, and STDP previously described, we incorporate a method of parameter dependence for the weights, based on the notion of power-law weight dependence for STDP \cite{gutig_learning_2003}. A generalized rule for power-law parameter dependence, for updates to a parameter $\theta$, would be:
\begin{equation}\label{eq:powerlaw-param-dep}
    \Delta \theta = (\theta_\text{max} - \theta)^{\mu_+} \Delta \theta_+ + (\theta - \theta_\text{min})^{\mu_-} \Delta \theta_-
\end{equation}
where $\theta$ is bound by a maximum $\theta_\text{max}$ and minimum $\theta_\text{min}$ and the utilizes respective exponents $\mu_+$ and $\mu_-$ on the interval $[0, 1]$. This assumes each update has a potentiative (positive) component $\Delta \theta_+$ and depressive (negative) component $\Delta \theta_-$, as is the case for DS-STDP and DR-STDP. Although this could be used for bounding the synaptic delays, a clamping step would still necessary since delays cannot be negative and must be not exceed the length of the stored buffer. In our experiments, we apply power-law bounding to the synaptic weights only.

We additionally apply a normalization step to synaptic weights after the application of an STDP-driven update:
\begin{equation}\label{eq:weight-norm}
    \mathbf{W}_{ji} \leftarrow \bar{w} \frac{\mathbf{W}_{ji}}{\lVert \mathbf{W}_{j}\rVert_{p}}
\end{equation}
where the vector of weights from every input neuron to an excitatory neuron $j$ must have an $\ell_p$-norm of $\bar{w}$. In our experiments, we use the $\ell_1$-norm and only apply this step to synaptic weights; however, we remark that it could reasonably be applied to synaptic delays as well.

DS-STDP and STDP are trained using cumulative exponential spike traces to represent the convolved spike trains. For delayed connections, prior values of the spike traces are buffered and continuous-time delays interpolate, when necessary, by decaying the trace by the corresponding amount when sampling between time steps.

For each model size, we train over a different number of epochs: $20$ for $100$ neurons, $40$ for $225$ neurons, $60$ for $400$ neurons, $80$ for $625$ neurons, and $100$ for $900$ neurons. The three models with more than $900$ neurons were all trained for $120$ epochs. We use the best testing accuracy for a given epoch to simulate early halting under ideal conditions. The hyperparameters between DS-STDP and STDP are kept the same for weights, but hyperparameters for DR-STDP are configured to slow the training in order to compensate for the non-sparse updates. In addition, as STDP only updates synaptic weights, the synaptic delays for models trained using STDP are set to zero rather than the random initialization shown in Table \ref{tbl:general-hyperparams}.

\begin{table}[!t]
    \centering
    \caption{Common training hyperparameters.}
    \label{tbl:train-common-hyperparams}
    \begin{tabular}{lcc}
        \toprule
        Hyperparameter & Value \\
        \midrule
        Weight Minimum ($w_\text{min}$) & $0$ \\
        Weight Maximum ($w_\text{max}$) & $1$ \\
        Weight Dependence Power ($\mu_+, \mu_-$) & $(1, 1)$ \\
        Per-Output L1-Norm of Weights ($\bar{w}$) & $78.4$ \\
        \midrule
        Delay Minimum ($d_\text{min}$) & $0 \text{ ms}$ \\
        Delay Maximum ($d_\text{max}$) & $10 \text{ ms}$ \\
        \bottomrule
    \end{tabular}
\end{table}

\begin{table}[!t]
    \centering
    \caption{Method-specific training hyperparameters.}
    \label{tbl:train-spec-hyperparameters}
    \begin{tabular}{clccc}
        \toprule
        & Trace Hyperparameters & DS-STDP & DR-STDP & STDP \\
        \midrule
        \multirow{4}{*}{\rotatebox{90}{Weights}}
        & Presynaptic Amplitude ($A_+$) & $5 \times 10^{-4}$ & $2.5 \times 10^{-4}$ & $5 \times 10^{-4}$ \\
        & Postsynaptic Amplitude ($A_-$) & $-5 \times 10^{-6}$ & $-2.5 \times 10^{-6}$ & $-5 \times 10^{-6}$ \\
        & Presynaptic Time Constant ($\tau_+$) & $20 \text{ ms}$ & $10 \text{ ms}$ & $20 \text{ ms}$ \\
        & Postsynaptic Time Constant ($\tau_-$) & $20 \text{ ms}$ & $10 \text{ ms}$ & $20 \text{ ms}$ \\
        \midrule
        \multirow{4}{*}{\rotatebox{90}{Delays}}
        & Presynaptic Amplitude ($A^\prime_-$) & $-1.2 \times 10^{-2}$ & $-6 \times 10^{-3}$ & --- \\
        & Postsynaptic Amplitude ($A^\prime_+$) & $1.2 \times 10^{-4}$ & $6 \times 10^{-5}$ & --- \\
        & Presynaptic Time Constant ($\tau^\prime_-$) & $20 \text{ ms}$ & $10 \text{ ms}$ & --- \\
        & Postsynaptic Time Constant ($\tau^\prime_+$) & $20 \text{ ms}$ & $10 \text{ ms}$ & --- \\
        \bottomrule
    \end{tabular}
\end{table}

\subsubsection{Simulation of Neurons and Synapses}
\label{sssec:neur-syn-sim}
We simulate the excitatory neurons using the adaptive leaky integrate-and-fire (ALIF) model \cite{chacron_interspike_2003} and the inhibitory neurons using the leaky integrate-and-fire (LIF) model \cite{stein_theoretical_1965}. Both models use the same dynamics, as given by the following equations:
\begin{align}
    \tau_m \frac{dV_m}{dt} &= - \left[V_m(t) - E_L\right] + R_m I(t) \label{eq:lif-membrane}\\
    V_m &\leftarrow V_\text{reset} &\text{if } V_m \geq \vartheta(t) . \label{eq:lif-reset}
\end{align}
The above specifies the change in the membrane potential $V_m$ as a function of the rest potential $E_L$, membrane time constant $\tau_m$, membrane resistance $R_m$, and input current $I$. After the membrane voltage is updated, assuming the neuron does not spike, Equation \ref{eq:lif-reset} specifies the spiking condition, i.e., when the membrane voltage exceeds a threshold voltage $\vartheta(t)$. For LIF neurons, this threshold voltage is always equal to some equilibrium value $\vartheta_\infty$. In contrast, ALIF neurons allow for the threshold to adapt based on spiking behavior. This adaptation is specified as follows:
\begin{equation}\label{eq:alif-thresh}
    \frac{d\vartheta}{dt} = - \frac{\vartheta(t) - \vartheta_\infty}{\tau_\vartheta} + \delta_\vartheta S(t) ,
\end{equation}
where the threshold exponentially decays back to its equilibrium with time constant $\tau_\vartheta$ and increases by $\delta_\vartheta$ whenever the neuron spikes.

The difference equations used to simulate the neuronal dynamics in our discrete simulation are derived directly from the exact solutions to the above differential equations. We use current-based synapses---those where the input current is independent of the membrane voltage of the post-synaptic neuron---modelled as delta synapses \cite{rosenbaum_modeling_2024}. This model is concretely defined as:
\begin{equation}\label{eq:delta-synapse}
    J(t) = Q \sum_{f \in \mathcal{F}} \delta_{\Delta t}(t - t^{(f)})
\end{equation}
\begin{equation}\label{eq:pulse-func}
    \delta_{\Delta t}(\tau) =
    \begin{cases}
        0 &\text{if } \tau < 0\\
        \frac{1}{\Delta t} & 0 \leq \tau < \Delta t \\
        0 &\text{if } \tau \geq \Delta t
    \end{cases}
\end{equation}
where $Q$ is the electric charge carried by each spike $f$ in the set of spikes $\mathcal{F}$ and $\delta_{\Delta t}$ is the pulse function with time step $\Delta t$. These synaptic currents are then modulated by weights and delays for integration into each neuron as follows:
\begin{equation}\label{eq:syn-current}
    I_j(t) = \sum_i \mathbf{W}_{ji} J_i(t - \mathbf{D}_{ji}) ,
\end{equation}
where $j$ is a post-synaptic neuron and each $i$ is a pre-synaptic neuron.

\begin{table}[!t]
    \centering
    \caption{Neuron and synapse hyperparameters.}
    \label{tbl:neuron-hyperparams}
    \begin{tabular}{lcc}
        \toprule
        Hyperparameter & Excitatory & Inhibitory \\
        \midrule
        Rest Potential ($E_L$) & $-65 \text{ mV}$ & $-60 \text{ mV}$ \\
        Reset Potential ($V_\text{reset}$) & $-60 \text{ mV}$ & $-45 \text{ mV}$ \\
        Membrane Time Constant ($\tau_m$) & $100 \text{ ms}$ & $75 \text{ ms}$ \\
        Membrane Resistance ($R_m$) & $1 \text{ M$\Omega$}$ & $1 \text{ M$\Omega$}$ \\
        Threshold Equilibrium ($\vartheta_\infty$) & $-52 \text{ mV}$ & $-40 \text{ mV}$ \\
        After-Spike Threshold Increment ($\delta_\vartheta$) & $5 \times 10^{-2} \text{ mV}$ & --- \\
        Threshold Time Constant ($\tau_\vartheta$) & $1 \times 10^{7} \text{ ms}$ & --- \\
        Refractory Period ($t_\text{ref}$) & $5 \text{ ms}$ & $2 \text{ ms}$ \\
        Spike Charge ($Q$) & $100 \text{ pC}$ & $75 \text{ pC}$ \\
        \bottomrule
    \end{tabular}
\end{table}

\section{Simulation Results}
\label{sec:sim-results}
At each tested model size, DS-STDP outperformed DR-STDP and STDP using both spike rate and spike responsiveness as the score on which the classification was performed: the specific results for which are given in Tables \ref{tbl:testacc-rate} and \ref{tbl:testacc-responsiveness}, respectively. The extended results for larger models trained using STDP are given in Table \ref{tbl:testacc-stdp-extended}. We also observed that DS-STDP required the most training epochs in order to achieve its best classification accuracy and that using responsiveness, rather than rate, as the scoring function for classification required more training epochs.

\begin{table}[!t]
    \centering
    \caption{Testing accuracy using spike rate.}
    \label{tbl:testacc-rate}
    \begin{tabular}{ccccccc}
        \toprule
        \multirow{2}{*}{Neurons} & \multicolumn{2}{c}{DS-STDP} &\multicolumn{2}{c}{DR-STDP} & \multicolumn{2}{c}{STDP} \\
        & Accuracy & Epoch & Accuracy & Epoch & Accuracy & Epoch \\
        \midrule
        100 & \textbf{0.8626} & (14) & 0.8415 & (10) & 0.8561 & (12) \\
        225 & \textbf{0.8836} & (24) & 0.8752 & (13) & 0.8773 & (17) \\
        400 & \textbf{0.8986} & (46) & 0.8919 & (21) & 0.8911 & (22) \\
        625 & \textbf{0.9104} & (78) & 0.8987 & (24) & 0.8943 & (18) \\
        900 & \textbf{0.9174} & (78) & 0.9011 & (25) & 0.8997 & (28) \\
        \bottomrule
    \end{tabular}
\end{table}

\begin{table}[!t]
    \centering
    \caption{Testing accuracy using spike responsiveness.}
    \label{tbl:testacc-responsiveness}
    \begin{tabular}{ccccccc}
        \toprule
        \multirow{2}{*}{Neurons} & \multicolumn{2}{c}{DS-STDP} &\multicolumn{2}{c}{DR-STDP} & \multicolumn{2}{c}{STDP} \\
        & Accuracy & Epoch & Accuracy & Epoch & Accuracy & Epoch \\
        \midrule
        100 & \textbf{0.8579} & (18) & 0.8426 & (20) & 0.8511 & (17) \\
        225 & \textbf{0.8868} & (37) & 0.8784 & (28) & 0.8734 & (22) \\
        400 & \textbf{0.9017} & (53) & 0.8937 & (44) & 0.8847 & (26) \\
        625 & \textbf{0.9119} & (78) & 0.8981 & (46) & 0.8928 & (40) \\
        900 & \textbf{0.9196} & (99) & 0.8982 & (56) & 0.8970 & (52) \\
        \bottomrule
    \end{tabular}
\end{table}

\begin{table}[!t]
    \centering
    \caption{Testing accuracy on larger models trained using STDP.}
    \label{tbl:testacc-stdp-extended}
    \begin{tabular}{ccccccc}
        \toprule
        \multirow{2}{*}{Neurons} & \multicolumn{2}{c}{STDP (Rate)} &\multicolumn{2}{c}{STDP (Responsiveness)} \\
        & Accuracy & Epoch & Accuracy & Epoch \\
        \midrule
        1225 & \textbf{0.9063} & (28) & 0.9031 & (47) \\
        1600 & 0.9001 & (41) & \textbf{0.9008} & (78) \\
        2025 & \textbf{0.9050} & (86) & 0.9040 & (72) \\
        \bottomrule
    \end{tabular}
\end{table}

\begin{figure}[!ht]
    \centering
    \includegraphics[width=0.65\linewidth]{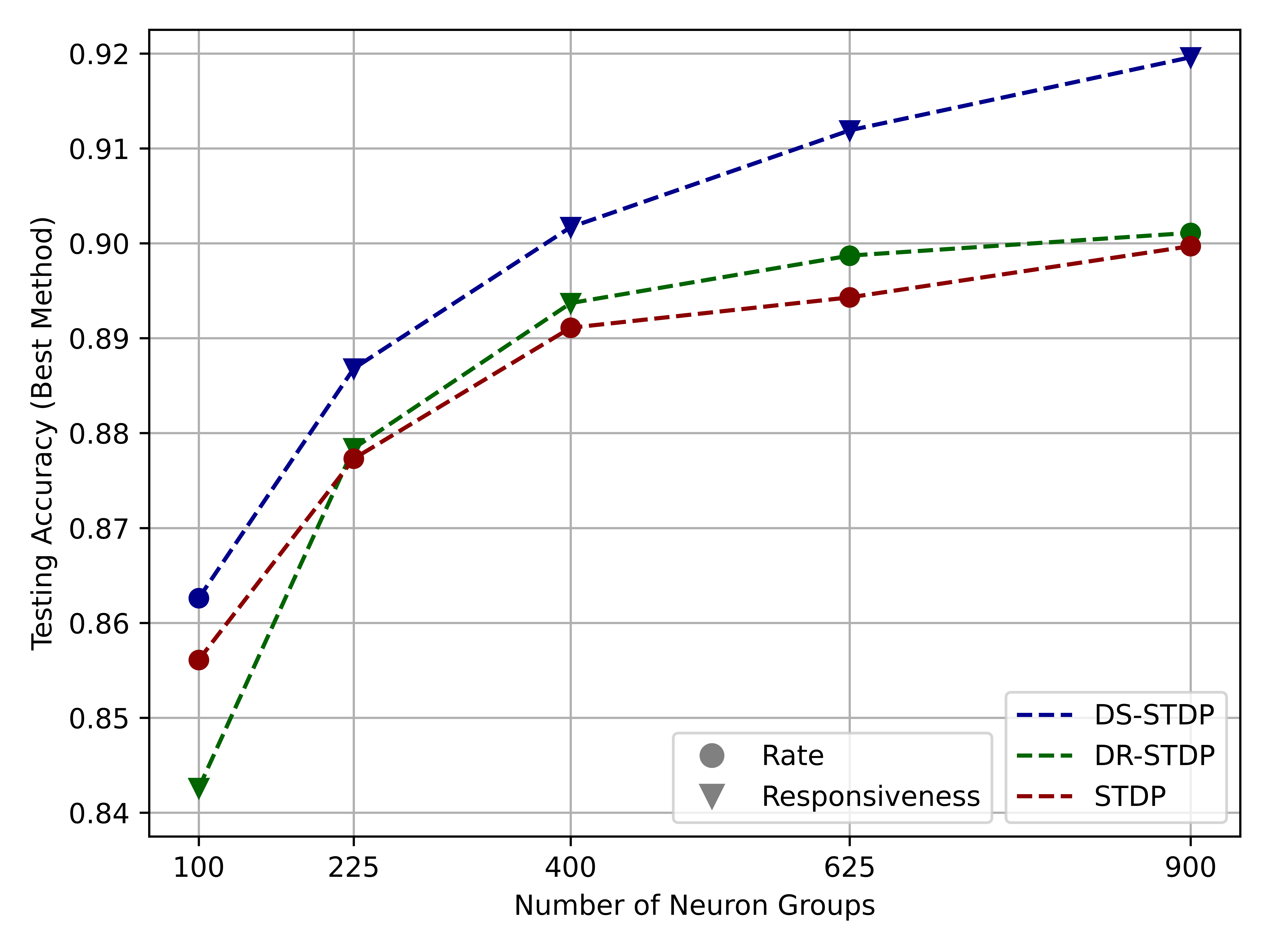}
    \caption{Testing accuracy using the better performing method for each configuration.}
    \label{fig:test-accuracy}
\end{figure}

\begin{figure}[!ht]
    \centering
    \begin{subfigure}[t]{0.4875\linewidth}
        \centering
        \includegraphics[width=\linewidth]{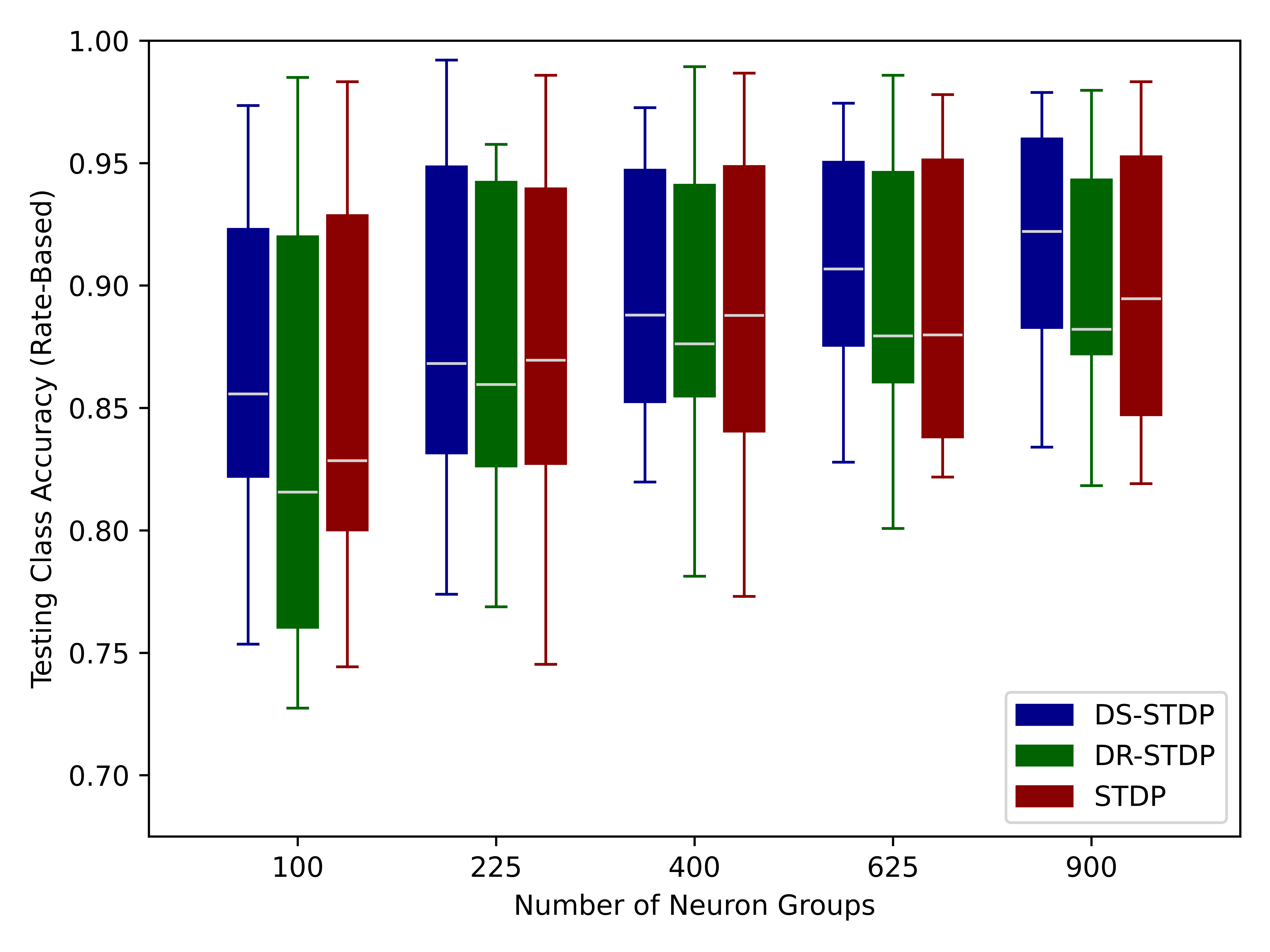}
    \end{subfigure}
    \hfill
    \begin{subfigure}[t]{0.4875\linewidth}
        \centering
        \includegraphics[width=\linewidth]{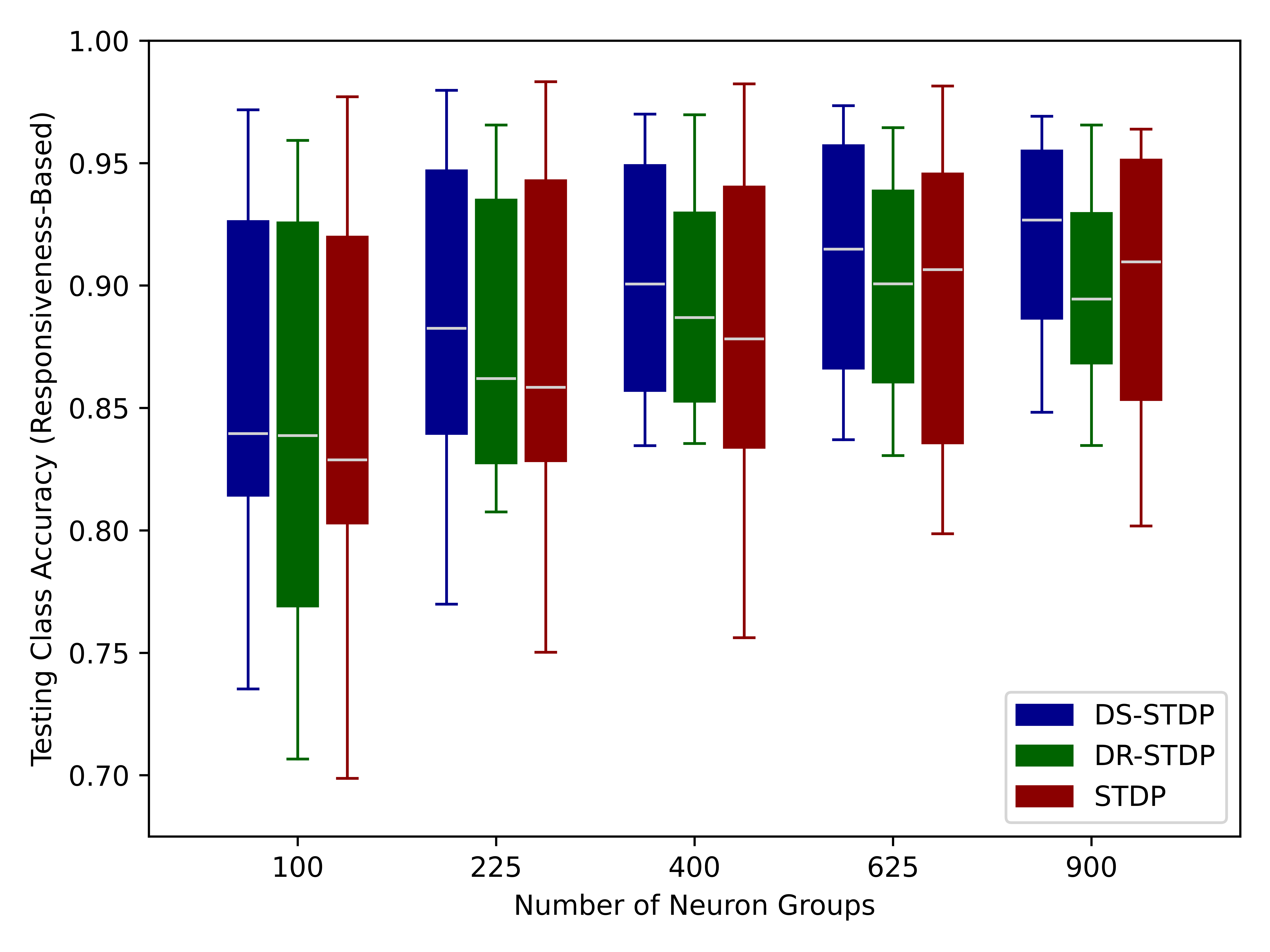}
    \end{subfigure}
    \caption{Five number summaries of per-class testing accuracy using spike rate (left) and spike responsiveness (right).}
    \label{fig:test-class-accuracy}
\end{figure}

Notably, we found that both methods incorporating trainable delays outperformed STDP in all but the smallest model, where only DR-STDP underperformed STDP, as is shown in Figure \ref{fig:test-accuracy}. From this result, we also observe that the increase in accuracy between the $625$ and $900$ neuron models, the two largest on which all three methods were tested, was greatest for DS-STDP ($0.0077$), versus DR-STDP ($0.0024$) and STDP ($0.0054$), suggesting that it has the most headroom for increasing performance with larger model sizes. The increased performance of DS-STDP holds when considering per-class testing accuracy across configurations. Figure \ref{fig:test-class-accuracy} shows the minimum, first quartile, median, third quartile, and maximum of per-class accuracies on the testing set. For all but two model sizes, DS-STDP maintained higher minimum class accuracies than DR-STDP or STDP.

\begin{figure}[!ht]
    \centering
    \includegraphics[width=0.65\linewidth]{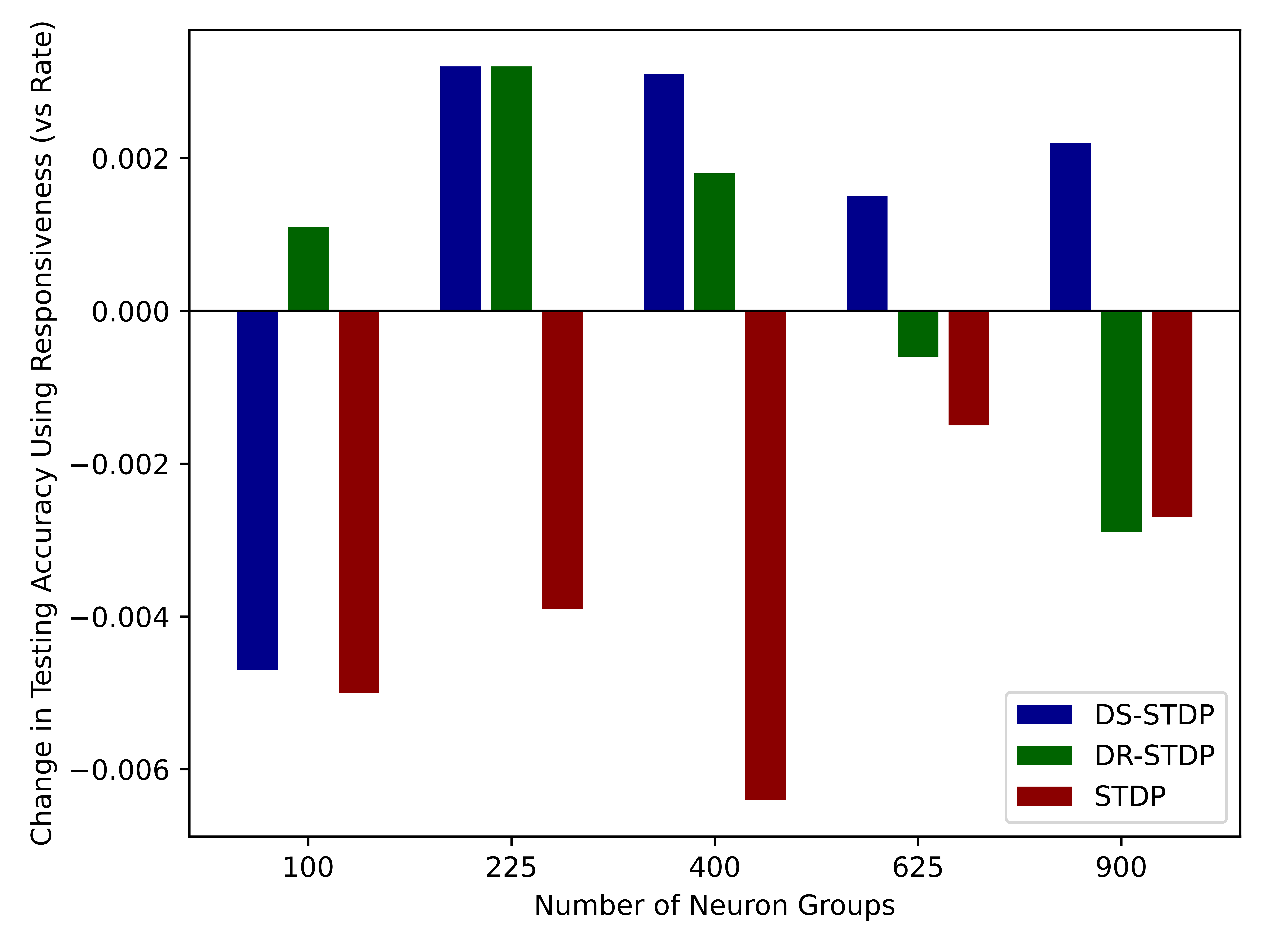}
    \caption{Difference in testing accuracy measurements between using spike responsiveness and spike rate. Positive values indicate spike responsiveness was more accurate.}
    \label{fig:accuracy-comparison}
\end{figure}

We found that the incorporation of trainable delays impacted whether spike rate or spike responsiveness was more performant as a scoring function used for classification. As shown in Figure \ref{fig:accuracy-comparison}, without trainable delays, spike rate was always the better scoring function to use (in the extended testing results, spike responsiveness slightly outperformed spike rate in one model). However, with the incorporation of trainable delays, spike responsiveness demonstrated superior performance in three out of five model sizes for DR-STDP (all but the two largest) and four out of five model sizes for DS-STDP (all but the one smallest). This suggests that models with trainable delays impart more information in spike responsiveness than models without delays, and furthermore, that both DS-STDP and DR-STDP impart this information to some degree. Additionally, this advantage persists with DS-STDP as the model scales up, whereas with DR-STDP it diminishes.

\begin{figure}[!ht]
    \centering
    \includegraphics[width=0.65\linewidth]{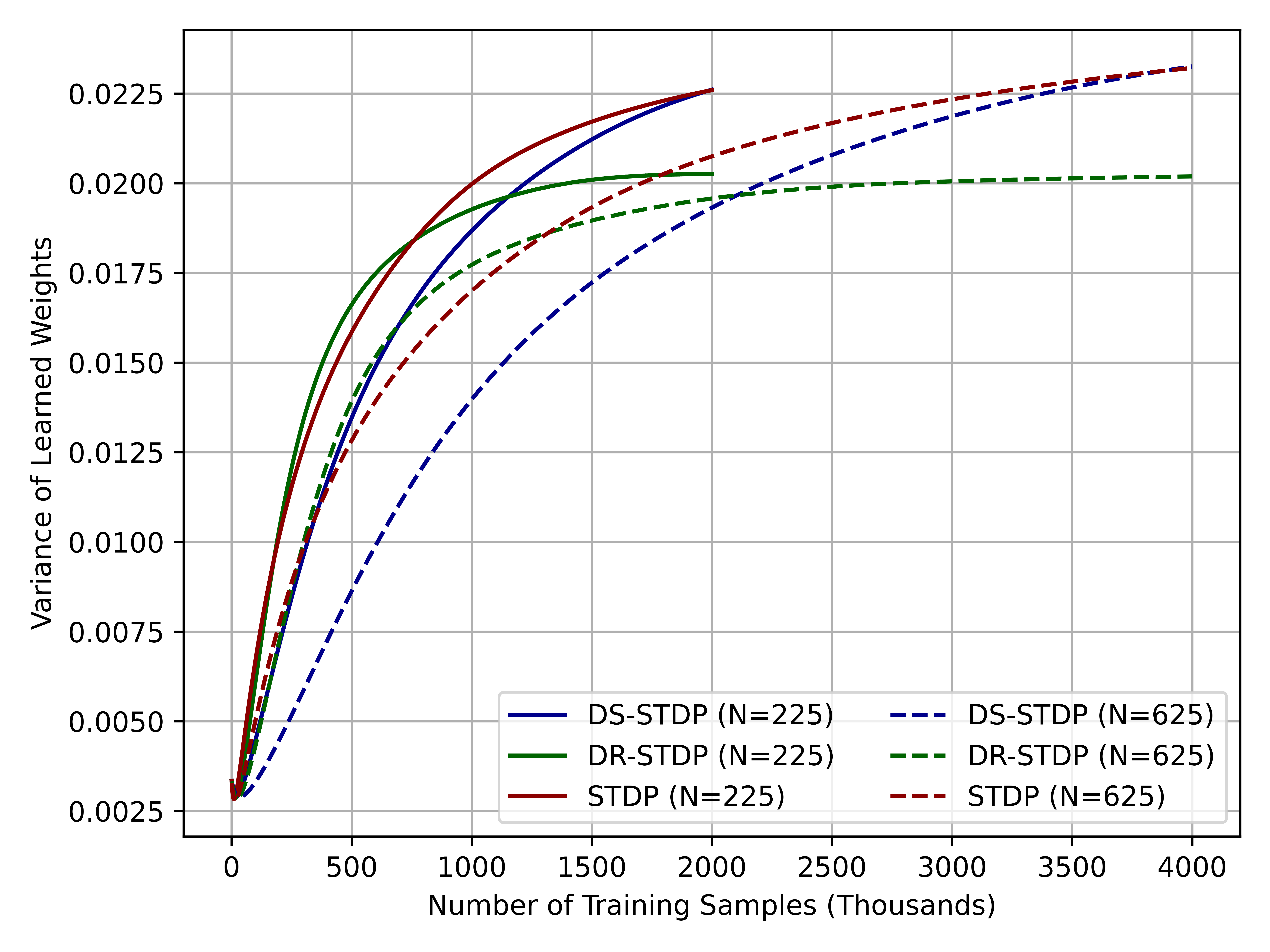}
    \caption{Variance in the learned weights during training.}
    \label{fig:weight-variance}
\end{figure}

As a proxy for training progress that is independent of the objective, we considered the variance of the learned connection weights. This is given in Figure \ref{fig:weight-variance} for models with $225$ and $625$ neuron pairs, although the patterns shown hold across all model sizes. The shape of the variance curves for models trained with DS-STDP and STDP are similar, whereas the shape for models trained with DR-STDP is significantly different. Examining the variance curves for DS-STDP and STDP, we see that although the progression in training between the two is similar, it occurs at a dampened rate for DS-STDP in spite of using identical weight update parameters. We attribute this to two possible causes: the first is that, under changing synaptic delays, the updates to synaptic weights are noisier relative to the optimal update; the second is that the increased model capacity, introduced via the synaptic delays, slows the training. In general, the larger the capacity of a model, the more training it will require. This can be seen here as well, where larger models trained with the same method required more training iterations in order to achieve the same variance in weights.

Although of the dampened rate for DS-STDP slowed the increase in variance, by the final training iteration, the variance in the model-learned weights was approximately the same as that of STDP. Meanwhile, DR-STDP had a sharper rise in spite of the reduced trace amplitude and time constant, and then it levels off sooner to a lower variance in the learned weights. Overall, this demonstrates that the learning dynamics of DS-STDP more closely resemble those of STDP than the learning dynamics of DR-STDP do.

\begin{figure}[!ht]
    \centering
    \includegraphics[width=0.65\linewidth]{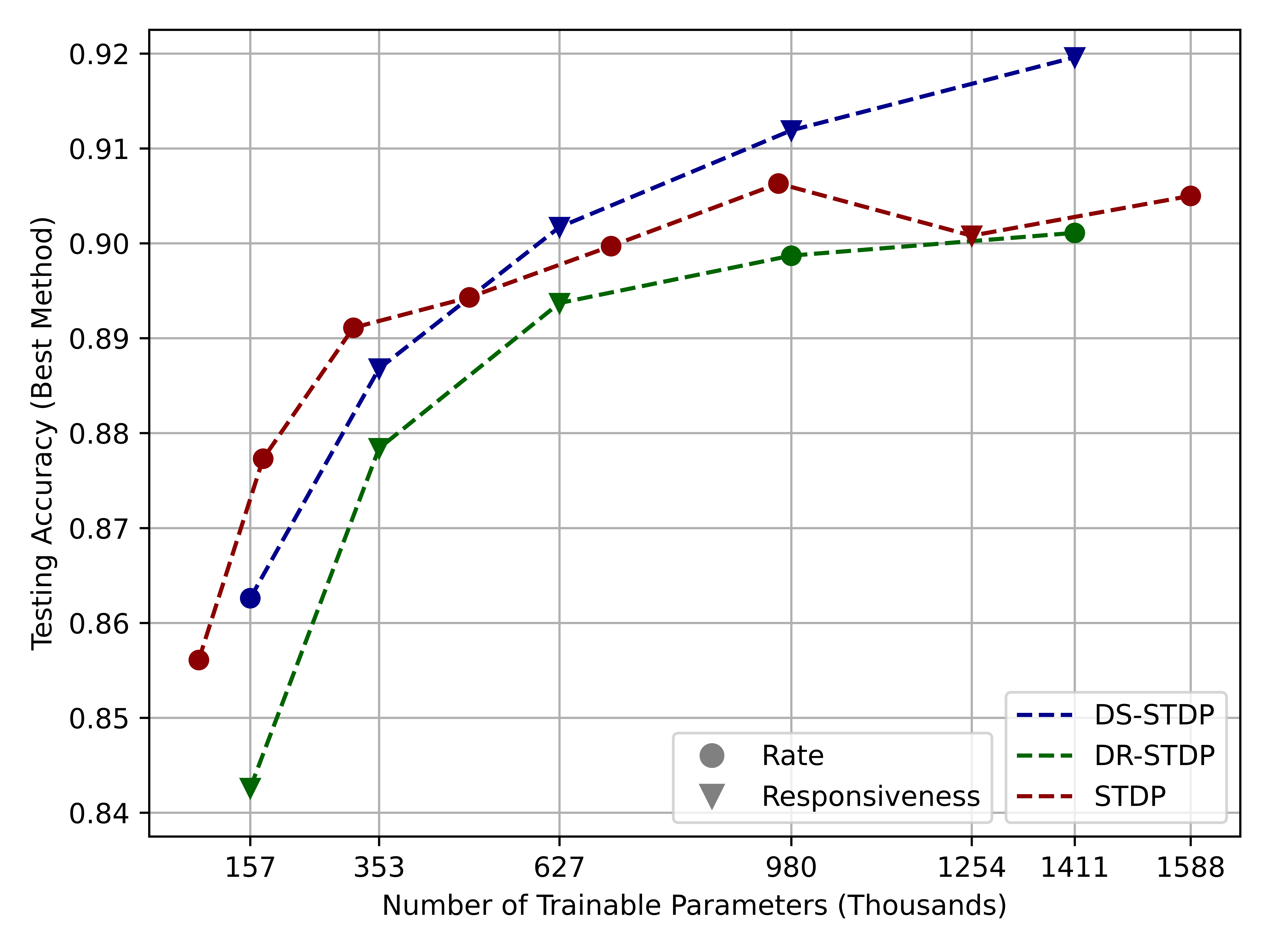}
    \caption{Testing accuracy when utilizing the better-performing method for each configuration, where the $x$-axis is scaled by the number of trainable weights and delays.}
    \label{fig:test-accuracy-paramscaled}
\end{figure}

While our primary focus was on developing a method for co-learning synaptic weights and delays, we can also extract some insight into the utility of synaptic delays. Figure \ref{fig:test-accuracy-paramscaled} shows the test accuracies but scales the $x$-axis according to the number of trainable parameters, excluding those parameters in the classifier and the adaptive thresholds of the excitatory neurons. This ``disadvantages'' the delay learning methods since, for a given number of trainable parameters, the classifier will only have access to half the number of neurons than would a model without trainable delays.

Even so, the $400$ neuron model trained with DS-STDP (627,200 trainable parameters) outperforms the $900$ neuron model trained with STDP (705,600 trainable parameters). We found that this trend continues when extending the testing and analysis to larger models trained with STDP. The $625$ neuron model trained with DS-STDP (980,000 parameters) outperformed every model trained with STDP, including the largest we tested with $2025$ neurons (1,587,600 parameters). This suggests that, since increasing the capacity of a model with only synaptic weights begins to yield diminishing returns, incorporating synaptic delays can provide a supra-additive increase in performance. That is, the interplay between synaptic weights and delays appears to meaningfully increase the model capacity itself.

\section{Conclusion}
\label{sec:conclusion}

In this work, we introduced delay-shifted spike-timing dependent plasticity \emph{DS-STDP}, a new method for the co-learning synaptic weights and synaptic delays for SNNs. We then provided an analytical comparison between DS-STDP and delay-related STDP (DR-STDP). In order to empirically compare them, we extended the Diehl \& Cook model to incorporate trainable synaptic delays. We additionally developed \emph{spike responsiveness}, a scoring function based on the time-to-first-spike that can serve as a drop-in replacement for spike rate based methods that require a scalar ``higher-is-better'' per-neuron score.

We then tested and compared the performance of DS-STDP to DR-STDP and STDP for classification on the MNIST dataset. From this, we found that, for a given number of neurons, DS-STDP consistently outperformed DR-STDP and STDP. Additionally, we demonstrated that, although using spike rate was better when synaptic delays were excluded, our spike responsiveness score generally yielded higher classification accuracies when trainable delays were incorporated into the model. This suggests that the latency information from spike trains is more effectively utilized when trainable delays are used.

More generally, this work also yields insights into synaptic delays. We found that the incorporation of trainable delays changes the training dynamics for the synaptic connection strengths. Although further research is required to confirm and understand this phenomenon, this study shows that, under some conditions, a model using trainable weights and delays may have a greater capacity than an equivalent model without synaptic delays but double the number of synaptic weights. Considering the known importance of synaptic delays in biological neural networks, better understanding of the relationship between synaptic weights and delays may prove valuable in understanding how biological brains learn as well as offer useful tools for brain-inspired computing and biomimetic intelligence \cite{ororbia_mortal_2024}.

\section{Data Availability}
\label{sec:data-availability}

All source code used to generate the results in this paper is available at \url{https://github.com/mdominijanni/dsstdp-results}. The simulation of spiking neural networks was performed using Inferno, a frozen version of which is included in the repository.

\printbibliography
\end{document}